\pdfoutput=1
\documentclass[10pt,twocolumn,letterpaper]{article}

\usepackage[dvipsnames]{xcolor}
\usepackage{iccv}
\usepackage{times}
\usepackage{epsfig}
\usepackage{amsmath}
\usepackage{amssymb}
\usepackage{booktabs} 

\input{header}
\usepackage{graphicx}

\usepackage[breaklinks=true,bookmarks=false]{hyperref} 
\iccvfinalcopy 

\def\iccvPaperID{2176} 

\ificcvfinal\fi

\usepackage[normalem]{ulem}
\usepackage{soul}
\usepackage{xstring}
\usepackage{xspace}

\sethlcolor{magenta}
\makeatletter
\def\SOUL@hlpreamble{%
    \setul{\dp\strutbox}{\dimexpr\ht\strutbox+\dp\strutbox\relax}%
    \let\SOUL@stcolor\SOUL@hlcolor
    \SOUL@stpreamble
}
\makeatother

\newcommand{\addref}[1]{\textcolor{magenta}{ [REF] }}

\DeclareMathOperator*{\argmax}{argmax}

\DeclareMathOperator{\cL}{\mathcal{L}}
\DeclareMathOperator{\cF}{\mathcal{F}}
\DeclareMathOperator{\cFb}{\overline{\mathcal{F}}}

\begin{document}

\title{Sparse-shot Learning with Exclusive Cross-Entropy \\for Extremely Many Localisations}

\author{Andreas Panteli$^{1, 2}$, Jonas Teuwen$^{1,2, 3}$, Hugo Horlings$^{1}$ and Efstratios Gavves$^{2, 4}$\\
$^{1}$Netherlands Cancer Institute, $^{2}$University of Amsterdam, \\$^{3}$Radboud University Medical Center, $^{4}$Ellogon.AI\\
{\tt\small \{a.panteli, j.teuwen, h.horlings\}@nki.nl, egavves@uva.nl}}

\maketitle
\ificcvfinal\thispagestyle{empty}\fi

\begin{abstract}
Object localisation, in the context of regular images, often depicts objects like people or cars.
In these images, there is typically a relatively small number of objects per class, which usually is manageable to annotate.
However, outside the setting of regular images, we are often confronted with a different situation.
In computational pathology, digitised tissue sections are extremely large images, whose dimensions quickly exceed 250'000 $\times$ 250'000 pixels, where relevant objects, such as tumour cells or lymphocytes can quickly number in the millions.
Annotating them all is practically impossible and annotating sparsely a few, out of many more, is the only possibility.
Unfortunately, learning from sparse annotations, or \emph{sparse-shot learning}, clashes with standard supervised learning because what is not annotated is treated as a negative.
However, assigning negative labels to what are true positives leads to confusion in the gradients and biased learning.
To this end, we present \emph{exclusive cross-entropy}, which slows down the biased learning by examining the second-order loss derivatives in order to drop the loss terms corresponding to likely biased terms.
Experiments on nine datasets and two different localisation tasks, detection with YOLLO and segmentation with Unet, show that we obtain considerable improvements compared to cross-entropy or focal loss, while often reaching the best possible performance for the model with only 10-40\% of annotations. 
\end{abstract}


\section{Introduction}

\begin{figure}[!t]
	\centering
    \includegraphics[height=9.5em]{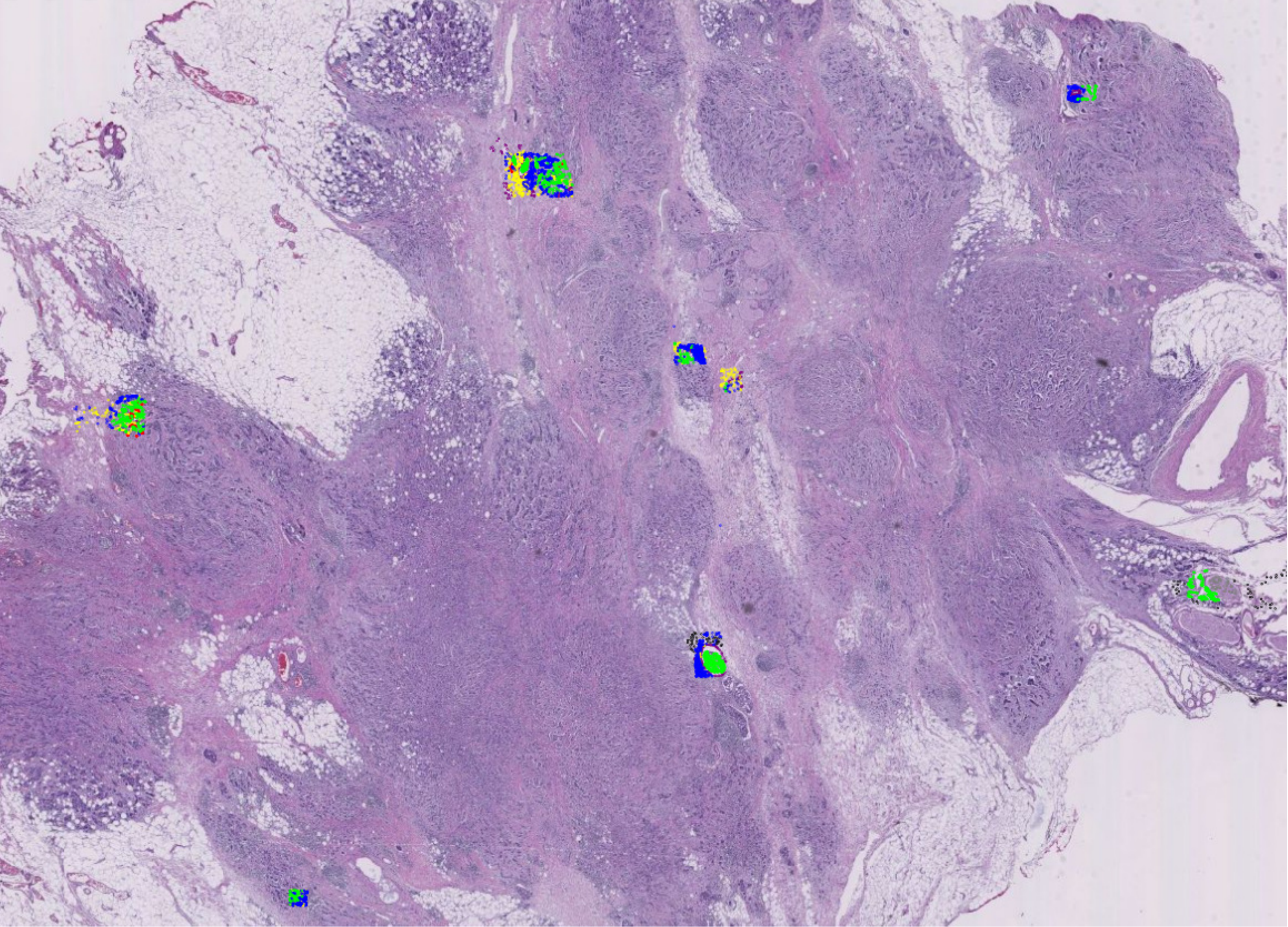}    
	\hfill
    \includegraphics[height=9.5em]{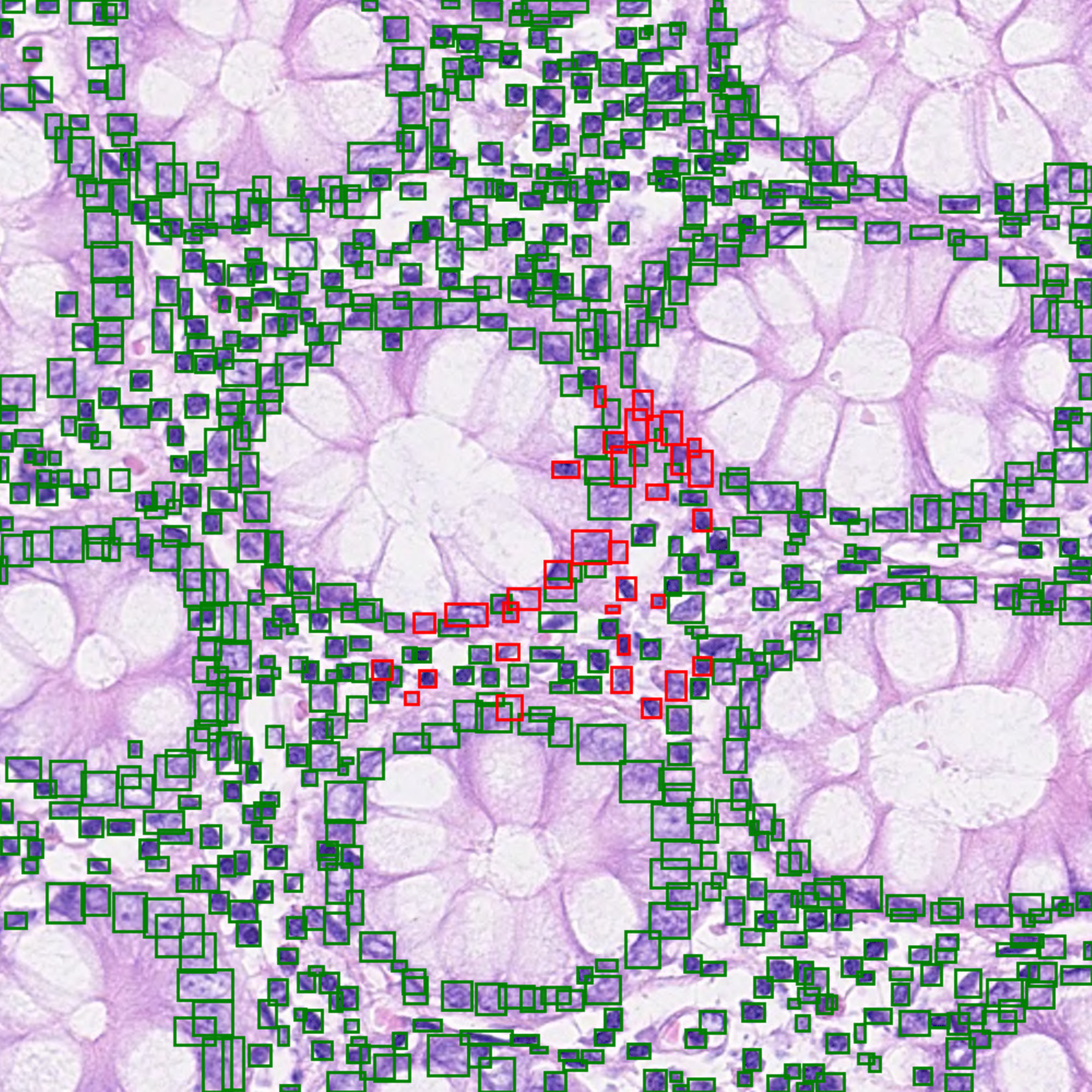}
	\caption{Left: A digitised tissue section containing millions of cells. An image typically corresponds to only a small part of smaller coloured-in regions in the whole slide. Only a handful of annotations are available, and after great effort (about 6'000 annotations in our data). Right: with red the non-exhaustively annotated objects in a small 1'000 by 1'000 region of a tissue slide image, roughly 30\% of the total number of objects in green (right).}
	\vspace{-1em}
	\label{fig:wsi_overview}
\end{figure}


With the advent of deep learning and big datasets, object localisation, be it bounding box detection \cite{tellez2018whole, lin2017focal, redmon2016you,van2018you, carion2020end, sirinukunwattana2016locality}, semantic segmentation \cite{unet,falk2019u,li2018h}, or instance segmentation \cite{mahbod2019two,graham2019mild,guerrero2018multiclass,graham2019hover}, has progressed with leaps and bounds ever since deformable part models \cite{girshick2015deformable,felzenszwalb2010cascade} and selective search \cite{uijlings2013selective,van2011segmentation}.
The basic assumption for all above localisation methods is that all relevant objects in the image are annotated.
This is a reasonable assumption for regular images like in PASCAL VOC 2007 \cite{everingham2015pascal} or MSCOCO \cite{lin2014microsoft}, containing on average 512 $\times$ 512 -or sometimes up to 1'000 $\times$ 1'000- images with no more than a dozen objects per class per image.
Outside the realm of regular images, however, we are often confronted with a different situation: digitised tissue sections are typically very large images, of file size around 1-10 GB, whose dimensions can quickly exceed 250'000 $\times$ 250'000 px, where relevant objects, such as tumour cells or lymphocytes can quickly number in the millions.
Annotating them all, even relying on regions-of-interest, can be hard and in practice only sparse annotations are feasible.
In this paper we focus on learning from sparse annotations, coined \emph{sparse-shot learning}, especially when the objective is localising an extreme numbers of objects. 

Learning from sparse annotations clashes with supervised learning, especially in the context of object localisation.
In the absence of any other knowledge, the typical assumption is to assign a negative label to all locations in the image that are not annotated as (true) positives.
This is a suboptimal choice on two grounds.
For one, it is very likely that the annotator could not annotate all relevant objects or that simply they missed many of them.
When blindly assuming as negative all unannotated areas, for example in digitised tissue sections, the unannotated objects often amount to more than 90\% of the total number of objects~\cite{hendry2017assessing, hendry2017assessing1, hendry2017assessing2}.
Secondly and more importantly though, assigning a negative label to what is in reality a true positive leads to conflicting gradients~\cite{shi2021marginal}, that in turn guide the model to poor convergence and generalisation.
Sparse-shot learning describes a setting, where standard supervised methods are ill-suited, both from a practical and methodological perspective \cite{saltz2018spatial}.

Learning given missing, or few, annotations has been explored in the past, albeit in different scenarios than sparse-shot learning.
In weakly supervised learning \cite{toussaint2018weakly, xu2019missing} an image-level label is provided, without localisation.
The model is then asked to jointly infer likely object locations, as well as learn an accurate classification model. However, when there exist no image-level label, as is the case in many object detection datasets, this type of weak learning cannot infer any localisation labels.
Some weakly supervised learning approaches include creating weak pseudo-labels for objects based on confident predictions \cite{misra2015watch, sohn2020simple}.
Sparse-shot learning is similar in that it assumes all unannotated areas to be potentially negatives, so in a way they correspond to weak negative labels.
A key difference is that sparse-shot learning focuses on rejecting specific subsets of these weak negative labels that are likely to add bias and, it does not create new positive labels for object detection.

Focusing on whole images rather than locations in images, in semi-supervised learning~\cite{xing2016robust} the goal is to learn from both annotated and unannotated images.
Unannotated images are hence leveraged to learn better and more general image-level classifiers.
Similarly, few-shot learning~\cite{yoon2019tapnet} utilises a small number of exhaustively annotated images.
Sparse-shot learning, on the other hand, describes a succinctly different setting often encountered in practice: learning localisation models from large images, where only a minute portion of the relevant locations are annotated during training.
In this work, our contributions are as follows:
\begin{enumerate}
    \item We introduce the problem of sparse-shot setting, which is predominant in several imaging scenarios in medical imaging, where acquiring high-quality exhaustive annotations is quite often downright impossible.
    \item We provide an analysis showing that the likely culprit leading to poor optimisations with sparse-shot learning is the high speed of learning attributed to biased annotations, and not the biased annotations themselves. To this end, we introduce a novel learning objective coined \emph{exclusive cross-entropy (ECE)} that incorporates a simple cut-off threshold to discard samples contributing large second-order derivatives to the loss, which are the ones speeding up biased learning.
    \item Via extensive experimentation on nine datasets and two state-of-the-art architectures, YOLLO \cite{van2018you} and Unet \cite{unet}, we show that the exclusive cross-entropy generalises in both detection and segmentation. Interestingly, the learned models trained in data, where only 10-40\% of the annotations are provided, often reach the same performance as the same models trained with exhaustive annotations, especially in segmentation tasks.
\end{enumerate}
\section{Related work}

Learning with weak supervision has been a popular area of research.
In the work of \cite{xu2019missing}, a weakly supervised learning method is proposed  to mediate the effect of partially annotated localisations.
They rely on a hybrid dataset containing both image- and instance-level labels, thus rendering the method applicable on data with only instance-level labels.
Recently, \cite{shi2021marginal} propose to use the similarity between classes (organs) to merge them together and train on a simpler, more general task.
Unlike our work, they rely on multiple classes that are similar while ignoring the background that is by definition dissimilar.

In the work of \cite{liu2020early}, noisy label learning was explored with loss regularisation focusing during the early stages of training. Similar to weak supervision, transfer learning is employed for the cases of only low information loss. In our sparse-shot learning, however, we have the extreme case of as little as 10\% of annotated objects. This results in noisy label learning, creating noisy pseudo-labels which accumulate biased gradient updates. In addition, early learning regularisation \cite{liu2020early} penalises the loss function based on the weak labels it iteratively creates, which can lead to a continuous cycle of wrong predictions caused by increasingly more mistakes.

Learning given outliers and imbalanced data has also been explored.
In the work of \cite{gupta2020robust} the Huber loss for dense object detection is used to address outlier samples.
The Huber loss aims to put smaller weights on outlier cases of hard examples that generate larger errors.
Non-exhaustive annotations, however, present themselves with a different challenge since the missing annotations are plentiful, and they are not outliers; down weighing them leads to discarding potentially important data during learning.
Recently, focal loss~\cite{lin2017focal} has also offered a significant step towards arbitrating the effect of unforeseen data imbalance, by using the model predictions to weigh more infrequent classes.
However, with non-exhaustive annotations the model predictions are inevitably biased due to the incorrect assignment of pseudo-labels to the unannotated data points.
Thus, focal loss is sensitive in the absence of exhaustive annotations.

\section{Sparse-shot learning}

We first introduce the problem setting of \emph{sparse-shot learning}.
We then discuss existing methods from the literature and present \emph{exclusive cross-entropy (ECE)}.

\subsection{Problem setting}

Let $I = (I_m)_{m=1}^M$ be a dataset of $M$ images, where each image $I_m = \{x_i, y_i\}, i \in [1, N]$ has a maximum of $N$ relevant objects, and each object, $x_i$, in the image is assigned one class $y_i \in \{1, \dots, C\}$ for $C$ number of classes.  To reduce notation clutter whenever the subscript $m$ can be inferred by the context, we drop it. $x_i$ can be a pixel, or a bounding box related to an object in an image.
For clarity of exposition, we focus first on the binary case, $y_i \in \{0, 1\}$.

In the standard fully supervised setting, a popular choice is the cross-entropy loss
\begin{align}
    \mathcal{L} = -\sum_{x_i \in I} \log p(y_i | x_i).
    \label{eq:cross-entropy}
\end{align}

For compactness, we will denote the positive predictions $p_i=p(y_i | x_i)$ and the negative ones by $1-p_i=1-p(y_i| x_i)$.

In sparse-shot learning, we \emph{do not have all} relevant labels at training time; that is, we do not have exhaustive knowledge of $y_i: \forall x_i \in I$.
Instead, we have the annotations $y_i$ for a few locations only, $x_i \in \cF$, where $\cF \subset I$ is our \emph{foreground} knowledge.
The rest of the unannotated image, $\cFb=I-\cF$, contains both irrelevant background (set $B$) $\cFb_B$ for which $y_i=0,~ \forall x_i \in \cFb_B$, as well as locations $\cFb_U$ that belong to one of the relevant classes, $y_i={1, .., C}$.
Expanding equation~\eqref{eq:cross-entropy} to incorporate these subsets, we have
\begin{equation}
    \begin{split}
        \mathcal{L} = &-\sum_{x_i \in \cF}  \log p_i \\
                    &-\sum_{x_i \in \cFb} \Big[ y_i \log p_i + (1-y_i) \log \big(1-p_i \big) \Big] \\
                    = & ~~\mathcal{L}_{\cF} + \mathcal{L}_{\cFb}
    \end{split}
    \label{eq:cross-entropy-sparse-shot}
\end{equation}

In the absence of any knowledge of annotations in $\cFb$, there exist two following options from the literature to compute the loss in equation~\eqref{eq:cross-entropy-sparse-shot}.

\paragraph{Unannotated regions as background.}
Following the paradigm of standard object localisation \cite{girshick2015fast,ren2015faster,he2017mask, lin2017focal}, all that is not included in the set of annotations is set to be background. That is, 
$\cFb \equiv B$.
This approach has the drawback that it includes true positive samples in the set of true negative samples, causing bias which adds to the loss as 
\begin{align}
    \text{bias}=-\sum_{x_i \in \bar{\mathcal{F}}_U} \log \big(1-p_i \big)
    \label{eq:bias-cross-entropy}
\end{align}
As a result, when optimising the parameters of the neural network, the model gets confused as it is asked to differentiate between samples that are virtually identical in appearance with opposite labels.
This pushes the model parameters to poor local minima and, thus, conflicting predictions.

\paragraph{Weak supervision.}
The predominant paradigm, in similar setups where annotations are partly missing from an image, is weakly supervised learning.
Many variants of weakly supervised learning have been explored \cite{vu2019methods, xu2019missing, diao2020dense, corredor2019spatial} in this context.
The general idea amongst them is that the model $f$ is trained for $R$ rounds.
The model from a previous round $t$ is used to predict the labels of unknown samples, $y_i = \argmax p(y_i | x_i; \theta_t)$, often referred to as \emph{pseudo-labels}.
The pseudo-labels are then used together with the true labels to minimise cross-entropy in equation~\eqref{eq:cross-entropy-sparse-shot} and obtain the updated model parameters, $\theta_{t+1}$. However, these new pseudo-labels introduce bias caused by wrong, false positive, assignments of objects, originally associated with the background set $\cFb$. These mistakes add a bias term to the loss described as 
%
\begin{align}
    \text{bias}=-\sum_{r \in R}\big[
        \sum_{x_i \in \bar{\mathcal{F}}_{U_{r,t}}} \log \big(1-p_i \big)
        + \sum_{x_i \in \bar{\mathcal{F}}_{r,t}} \log \big(1-p_i \big)
        \big]
    \label{eq:bias-weak-supervision}
\end{align}
where $\mathcal{F}_{U_{r,t}} \subseteq \mathcal{F}_U$, $\cF_{r,t}$ corresponds to the weakly annotated labels by model $t$ at round $r$. In that respect, weak supervision might eventually do more harm than good, because it biases the final classifier not only on one label side ($y_i=0$) but all.


\subsection{Motivation for exclusive cross-entropy} 

In the absence of exhaustive ground truth knowledge in the background, any learning algorithm will inevitably introduce bias to the model parameters.
Ideally, for sparse-shot learning, we want an algorithm that takes advantage of the background without disproportionately biasing the model parameters \emph{either towards pseudo-positive or pseudo-negative} labels.

To this end, rather than fixating on how to optimally infer the missing annotations $y_i: \forall x_i \in \cFb$, we focus on the learning dynamics of the classifier and how we can optimally influence these dynamics in the sparse-shot learning setting.
The objective is to discover background samples -positive or negative- that are likely to add significant bias to the learning, and skip them.
Specifically, in the absence of any knowledge of annotations in the background, we tentatively consider all samples in the background as negative samples such that we at least do not add bias to the positive samples in the training set; as noted in equation \eqref{eq:bias-cross-entropy}.

\begin{figure}[!t]
    \centering
    \includegraphics[width=0.8\linewidth]{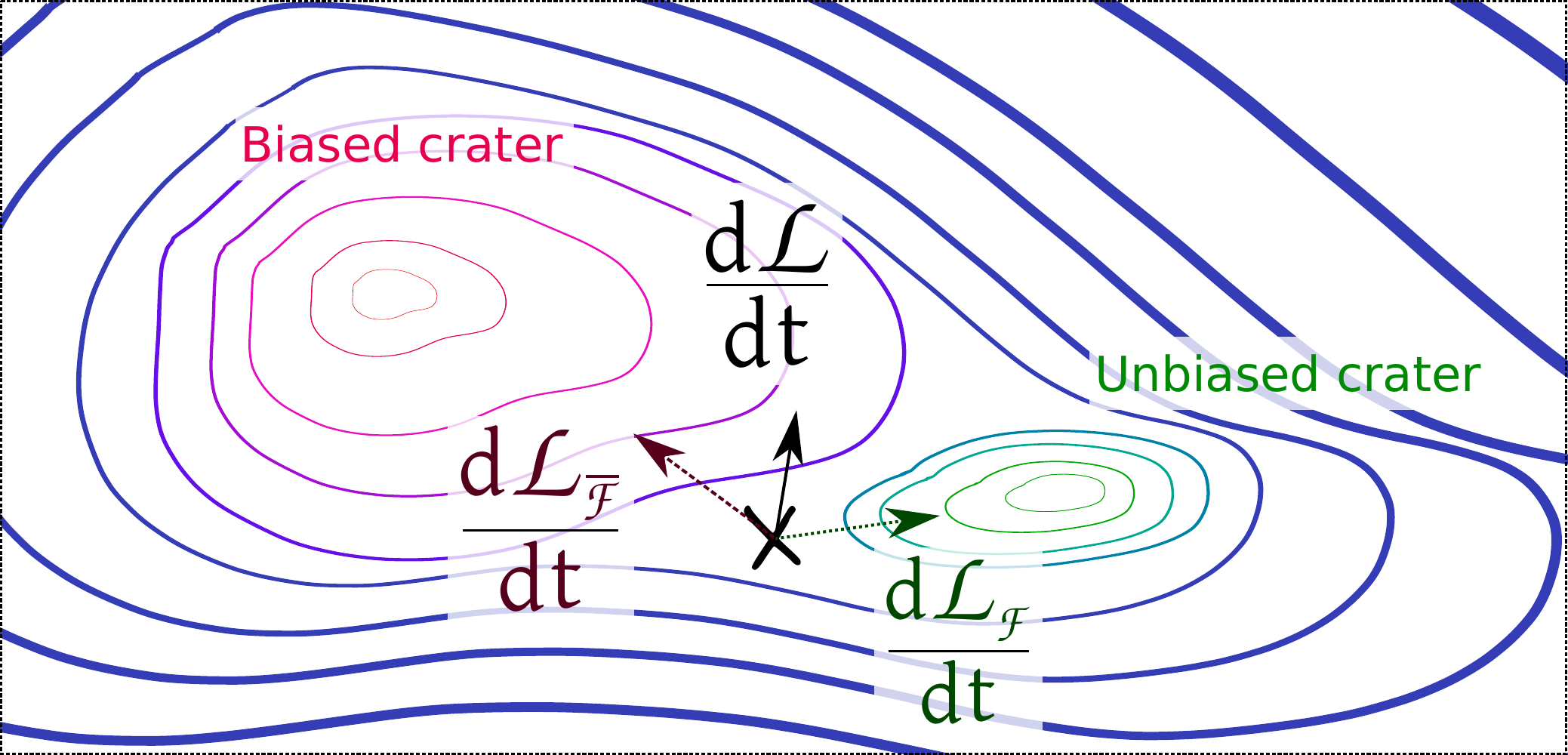}
    \caption{When annotations are missing, unannotated data cause learning models to near biased craters due to incorrect information. One way to improve, is to eliminate all instances containing bias, but that would mean eliminating all background examples in non-exhaustively annotated sets. Instead, we propose to simply slow down the pace of learning, captured by $\frac{d^2\cL}{dt^2}$, coming from the unannotated data while maintaining the pace of learning from certain annotated data. That way, the model will move faster towards the unbiased crater and eventually the desired solution in spite of the unavoidable bias.}
    \vspace{-1em}
    \label{fig:contour-gradient}
\end{figure}

\vspace{-1em}
\subsubsection{Exclusive cross-entropy for sparse-shot learning} 

Although in the beginning of the training, any prediction will likely be highly inaccurate, a model is still prone to returning highly confident predictions for both samples in $\cF$ and $\cFb$.
This is known as overconfidence in neural network predictions when the softmax and sigmoid activation functions are used for classification \cite{korteling2018neural, kristiadi2020being, mozejko2018inhibited}.
This effect is due to the nature of cross-entropy in equation~\eqref{eq:cross-entropy}, which attains the lowest score when the model predictions $\log p$ are the highest (either $p=1$ or $1-p=1$ for positive and negative samples respectively). 
It is particularly problematic in the case of missing annotations, as the model will be encouraged to make overconfident predictions for samples in the training, whose annotations are not given but inferred; thus often wrong. 

To motivate how to exploit the learning dynamics to break out of this paradox, we illustrate in figure \ref{fig:contour-gradient} a hypothetical optimisation landscape in gradient descent.
Figure \ref{fig:contour-gradient} highlights the scenario where the bias and the dynamics of learning may adversely affect the final solution.
For the purpose of the explanation and without loss of generality, in this example we assume we have one \emph{unbiased minimum} centred in an unbiased crater, which we would obtain if we had perfect knowledge of all relevant annotations in the background.
Next to our unbiased crater, there exist multiple biased ones caused by the addition of biased annotations.
In reality, neural networks exhibit multiple equivalent minima, however, this does not affect the motivation. Our hypothesis is that when learning from incorrect annotations, models converge to the minima in biased craters; \ie their performance would \emph{not} be as good as models trained on all correct annotations.

Ideally, we want the model to enter the unbiased crater, as in that case it will almost certainly converge to the optimal parameters with standard gradient descent.
Unfortunately, the biased gradients will inevitably push the model towards one of the biased craters.
One way to limit this, is to make sure the model learns from the unannotated background samples at a slower speed than it does from the certain foreground ones.
As learning is captured by the first derivative of the loss with respect to time, $\frac{d \cL}{dt}$ (derivatives with respect to parameters correspond to optimal model steps), the speed of learning is captured by the second derivative with respect to time, $\frac{d^2 \cL}{dt^2}$.
In other words, we want the second derivative of the background loss to be small, or even zero, compared to the second derivative of the foreground loss, \ie,
\begin{align}
    \frac{d^2 \cL_{\cFb}}{dt^2} \ll \frac{d^2 \cL_{\cF}}{dt^2}
    \label{eq:derivatives}
\end{align}
If equation~\eqref{eq:derivatives} holds, that indicates that the model learns faster from the positive samples, compared to negative ones, thus increasing the chances of reaching the unbiased crater before getting trapped in a biased one.
Moving the detailed computations to the supplementary material, the derivative equation can be expressed as
\begin{align}
    \frac{d^2 \cL_{\cFb}}{dt^2} \propto p^m (1-p)^n,
\end{align}
with polynomial roots $p=0$ and $1-p=0$.
To make sure that the second order derivative is zero or almost zero, we shall exclude training samples, in the unannotated areas $\cFb$, which have high confidence predictions. Since all unannotated samples are assigned a weak negative label, we introduce an exclusivity threshold term $\rho$ to the cross-entropy loss in equation~\eqref{eq:cross-entropy-sparse-shot}, only for the unannotated areas $\cFb$.
\begin{align}
    \mathcal{L} &= -\sum_{x_i \in \cF}  \log p_i
                   -\sum_{x_i \in \cFb} \delta(p_i<\rho^\beta) \log \big(1-p_i \big)
    \label{eq:exclusive-cross-entropy}
\end{align}
where $\beta$ is an annealing hyper-parameter and $\delta(\cdot)$ is the Kronecker delta function. As learning progresses and the model improves, predictions will become successively more confidently accurate and, hence, the threshold requirement can be relaxed over time by a less strict $\beta$. 
Note that equation~\eqref{eq:exclusive-cross-entropy} can support multiple classes by modifying the log probability $\log \big(1-p_i \big)$ accordingly.
We refer to the loss in equation~\eqref{eq:cross-entropy-sparse-shot} as \emph{exclusive cross-entropy} (ECE).

\subsubsection{Intuitive motivation and discussion}

Using exclusive cross-entropy enforces that the model should not be over-confident when it is too early for any model to be accurate. 
In contrast to cross-entropy, exclusive cross-entropy attempts to ignore risky high-confidence predictions and does not encourage the model to assign high scores to as many samples as possible.
High-confidence predictions, without sufficient training, run the risk of being false positives/negatives and will wrongfully push the model in the wrong direction.
Low-confidence unannotated data corresponding to false negatives (\ie unlabelled objects), on the other hand, will have small gradient magnitudes due to their low score, but their direction will, as learning progresses, hopefully tend to be in the right, approximate, direction. 
In order to avoid converging too quickly to spurious local minima, the goal is to slow down the learning speed from high-risk unannotated data and reach an unbiased crater first.

Specifically, in the beginning of standard training, the classifier is \emph{de facto} imprecise.
Any confidence in predictions are, thus, likely to be misplaced, certainly so for training samples that miss a manual annotation.
Given that our background training samples are all considered as tentative negatives ($y_i=0, \forall x_i \in \cFb$), let us consider the case of a high confidence positive prediction, $p(y_i=1 | x_i) > \rho$, for a real true positive object.
The first possible reasoning, is that the model is already capable of recognising objects correctly as positive predictions, $y_i=1$.
This means that the model is already accurate and there is no reason it should receive an update via back-propagation.
The second possibility, is that the pseudo-negative annotation is wrong. Back-propagating would update the model towards an incorrect direction.
Therefore, not only there is no big need to update the model, but we could be adding bias due to incorrect pseudo-annotations.
Given that we do not really know the true label, it is, therefore, better to exclude the contribution of this training sample to the gradient at this round.
A similar argument can be constructed for high confidence negative predictions, $p(y_i=0 | x_i) > \rho$.

It is important to note that the annealing and exclusivity threshold employed, are not equivalent as changing the learning rate nor ignoring unannotated objects altogether. Exclusive cross-entropy is similar to a dynamic switchable learning rate; where the rate is dynamically set to zero if training examples are unannotated.

\paragraph{Computational cost.}
As the exclusive cross-entropy is computed using the already calculated $p(y_i | x_i)$, the computational cost is virtually identical to standard cross-entropy.
No retraining, compared to weakly supervised learning, or other expensive processes are required.

\paragraph{Annealing $\rho$.}
Our primary objective when satisfying equation~\eqref{eq:derivatives} is that the model reaches the unbiased crater first.
Once in the unbiased crater, the model will eventually reach the desired minimum.
By annealing threshold $\rho$ by parameter $\beta$ we ensure that learning is influenced less by biased loss terms at the early stages and takes more samples into account at the later stages.
In experiments, we find that the learning algorithm is robust with respect to $\rho$ and $\beta$; so we use the same $\rho$ and $\beta$ for all our datasets and obtain consistently good performance.

\paragraph{Class imbalance.}
In object localisation, class imbalance can have strong effects on learning~\cite{redmon2016you, van2018you}.
Especially in large images such as the tissue sections, the amount of irrelevant or background object instances dwarf in comparison to the few positive annotations provided by the annotator.
To account for the severe class imbalance, we can complement the exclusive cross-entropy with the focal loss re-weighting scheme, $u(p_i) = -\alpha (1 - p_i)^\gamma \log(p_i)$, 
as originally proposed by \cite{lin2017focal}.
In this case, the \emph{focal} exclusive cross-entropy is computed as
\begin{align}
    \mathcal{L} &= -\sum_{x_i \in \cF}  \log p_i
                   -\sum_{x_i \in \cFb} \delta(p_i<\rho^\beta) u\big(1 - p_i \big)
    \label{eq:focal-exclusive-cross-entropy}
\end{align}

\section{Experiments}


\subsection{Experimental Setup}

\paragraph{Data.}
We evaluate on the following nine datasets: CoNSeP \cite{graham2019hover}, CPM15 \cite{vu2019methods}, CPM17 \cite{vu2019methods}, CRCHisto \cite{sirinukunwattana2016locality}, Kumar \cite{kumar2017dataset}, MoNuSeg \cite{kumar2017dataset}, WBC-NuClick \cite{koohbanani2020nuclick}, TNBC \cite{naylor2017nuclei}, and our own tumour-infiltrating lymphocyte (TIL) localisation benchmark containing 16 Hematoxylin and Eosin (H\&E) stained digital biopsies of whole slide images (WSIs).
The largest dataset is TIL with  440'734 images and 45'127 cell annotations, including the 6'631 lymphocytes.
The second-largest dataset is WBC-NuClick with 1'463 images, while the second most annotated dataset is the CRCHisto dataset with 29'748 cells. We provide all details and visual examples in the supplementary material.

\paragraph{Evaluation.}
%

All datasets, except for TIL, contain images that are only small portions of the digitised tissue sections, such that they can be exhaustively annotated.
We create non-exhaustive annotation set variants with 10\%, ..., 90\% of the annotations (100\% is the full set).
To make sure the different variants are comparable, we include all annotations in every smaller variant in the variants above (the 80\% variant annotations are also in the 90\% variant and so on).

We evaluate segmentation using DICE and object detection using F1 score. The TIL dataset contains only a very small portion of all cells, thus we cannot use precision-related metrics, as unknown true positives would be counted as true negatives.
Instead, given that in the TIL dataset we have annotations for other cell types that are similar to lymphocytes and are the most likely false positives, we propose the \emph{exclusive recall} computed as $\text{Rec}_{\text{exc}}(y)=\text{Rec}(y) \cdot (1-\text{Rec}(\neq y))$.
While still not accounting for the missed true positives, exclusive recall down weighs the score when predictions correspond to wrong cell types and can quantitatively score \emph{relative} performance between methods.

\paragraph{Architectures.}
The exclusive cross-entropy is agnostic to the specific segmentation or detection model and architecture.
We experiment with two state-of-the-art methods: YOLLO \cite{van2018you} for object detection and Unet \cite{unet} for segmentation using standard open-source implementation.
We train the models from scratch per non-exhaustive set (10 sets per dataset) and with no pre-training. 
For hyper-parameter tuning we rely only on the TNBC dataset for segmentation, and reuse the same parameters in all other datasets, experiments and tasks (both for segmentation and detection).
We use \emph{no task-specific or dataset-specific parameters} for the exclusive cross-entropy. 
In all 161 experiments with exclusive cross-entropy with YOLLO and Unet on all nine datasets use the same hyperparameter values, to demonstrate generality and robustness. We include in the supplementary material all the model and training parameters.

\subsection{Ablation study}

\paragraph{Cross-entropy variants and weakly supervised learning.}
We report results with exclusive cross-entropy as well as weakly-supervised learning on the 30\% and 60\% non-exhaustive variants of the TNBC dataset.
We start with training using standard cross-entropy.
Then, we use the trained model to update the labels in the respective non-exhaustively annotated training sets.
If the prediction for an unannotated sample is $p_i > \tau$, then the sample becomes a pseudo-positive, we re-train and repeat the process. Noisy label learning with early learning regularisation \cite{liu2020early}, performs similar to the standard weakly-supervised learning, as shown further in the supplementary material.

We present results in table \ref{tab:weak-supervision-results}.
We observe that weak supervised learning does not increase the performance of the standard cross-entropy training.
A possible reason -in contrast to the regular uses of weak supervision \cite{campanella2019clinical}- is that objects in medical images are easy to confuse.
Weak supervision works better when the expected confusion is not high.
For standard cross-entropy, focal reweighing is not beneficial, likely due to overly down-weighing the actual true positives.
Adding focal reweighing, to the unannotated $\cFb$ group, in the exclusive cross-entropy is beneficial and hence, we use focal reweighing in all subsequent experiments with exclusive cross-entropy.

\begin{table}
    \caption{Exclusive cross-entropy \emph{vs.} weak supervision for 30\% and 60\% annotations on TNBC for the detection task.}
    \vspace{-1em}
    \begin{center}
        \begin{small}
        \begin{tabular}{lccc}
            \toprule
               & $\tau$ & F1@30\% & F1@60\% \\
            \midrule
            Cross-entropy                          &        & 0.65 & 0.7 \\
            +weak supervision                      & $0.75$ & 0.64 & 0.68 \\
            +weak supervision                      & $0.50$ & 0.62 & 0.64 \\
            +focal loss                            &        & 0.36 & 0.49 \\
            \midrule
            Exclusive cross-entropy                &        &   0.70     & 0.75  \\
            +focal loss                            &        &\textbf{0.74}  & \textbf{0.80} \\
            \bottomrule
        \end{tabular}
        \end{small}
    \end{center}
    \vspace{-1em}
    \label{tab:weak-supervision-results}
\end{table}

\paragraph{Exclusivity threshold and annealing schedules.}
Next, we ablate different exclusivity thresholds and annealing schedules.
We present two experiments with fixed $\rho$ at $0.5$ and $0.75$ ($\rho=1$ is standard cross-entropy).
We also present two experiments with linear and sigmoid scheduling in annealing $\rho$.
We gather results in table \ref{tab:annealing-results}.
We observe consistently good performance no matter the type of threshold and scheduling, with sigmoid scheduling doing best.
In the following experiments, we will use the sigmoid schedule.


\begin{table}
    \caption{Annealing scheduling study on the 30\% and 60\% sets of TNBC for the detection task.}
    \vspace{-1em}
    \begin{center}
        \begin{small}
        \begin{tabular}{lcc}
            \toprule
                                                                     & F1@30\% & F1@60\% \\
            \midrule
            Fixed $\rho=0.75$                                        & 0.68 & 0.70 \\
            Fixed $\rho=0.50$                                        & 0.66 & 0.71 \\
            Linear $\rho= 0.75 \cdot\rho_t, \rho_t: 0 \rightarrow 1$ & 0.71 & 0.73 \\
            Sigmoid $\rho= \sigma(\rho_{Linear})$  & \textbf{0.74} & \textbf{0.80} \\
            \bottomrule
        \end{tabular}
        \end{small}
    \end{center}
    \vspace{-2em}
    \label{tab:annealing-results}
\end{table}

\subsection{Sparse-shot segmentation}
\begin{figure*}[t!]
    \centering
    \begin{subfigure}{0.24\linewidth}
        \begin{scaletikzpicturetowidth}{\textwidth}
            \begin{tikzpicture}[scale=\tikzscale]
                \begin{axis}[
                    title={TNBC},
                    xlabel={Percentage of cell annotations [\%]},
                    ylabel={DICE},
                    xmin=5, xmax=105,
                    xtick={10,20,30,40,50,60,70,80,90,100},
                    legend pos=south east,
                    ymajorgrids=true,
                    grid style=dashed,]
                \addplot
                    coordinates {(10,0.73)(20,0.77)(30,0.8)(40,0.82)(50,0.82)(60,0.83)(70,0.82)(80,0.83)(90,0.83)(100,0.82)}; \addlegendentry{ECE (ours)}
                \addplot
                    coordinates {(10,0.44)(20,0.43)(30,0.43)(40,0.51)(50,0.52)(60,0.61)(70,0.80)(80,0.80)(90,0.81)(100,0.84)};
                    \addlegendentry{CE}
                \addplot[dashed, color=cyan, mark=otimes*]
                    coordinates {(10,0.69)(20,0.64)(30,0.61)(40,0.63)(50,0.62)(60,0.64)(70,0.64)(80,0.64)(90,0.64)(100,0.64)};
                    \addlegendentry{FL}
                \addplot[dashed, color=brown, mark=square*]
                    coordinates {(10,0.44)(20,0.44)(30,0.43)(40,0.42)(50,0.42)(60,0.62)(70,0.77)(80,0.78)(90,0.82)(100,0.83)};
                    \addlegendentry{Huber}
                \end{axis}
            \end{tikzpicture}
        \end{scaletikzpicturetowidth}
    \end{subfigure}
    \hfill
    \begin{subfigure}{0.24\linewidth}
        \begin{scaletikzpicturetowidth}{\textwidth}
            \begin{tikzpicture}[scale=\tikzscale]
                \begin{axis}[
                    title={CoNSeP},
                    xlabel={Percentage of cell annotations [\%]},
                    ylabel={DICE},
                    xmin=5, xmax=105,
                    xtick={10,20,30,40,50,60,70,80,90,100},
                    legend pos=south east,
                    ymajorgrids=true,
                    grid style=dashed,]
                \addplot
                    coordinates {(10,0.69)(20,0.75)(30,0.78)(40,0.81)(50,0.80)(60,0.80)(70,0.8)(80,0.79)(90,0.8)(100,0.79)}; \addlegendentry{ECE (ours)}
                \addplot
                    coordinates {(10,0.49)(20,0.47)(30,0.47)(40,0.49)(50,0.61)(60,0.61)(70,0.72)(80,0.76)(90,0.8)(100,0.8)};
                    \addlegendentry{CE}
                \addplot[dashed, color=cyan, mark=otimes*]
                    coordinates {(10,0.68)(20,0.61)(30,0.61)(40,0.59)(50,0.61)(60,0.54)(70,0.63)(80,0.62)(90,0.63)(100,0.63)};
                    \addlegendentry{FL}
                \addplot[dashed, color=brown, mark=square*]
                    coordinates {(10,0.48)(20,0.46)(30,0.44)(40,0.49)(50,0.48)(60,0.56)(70,0.67)(80,0.73)(90,0.72)(100,0.72)};
                    \addlegendentry{Huber}
                \end{axis}
            \end{tikzpicture}
        \end{scaletikzpicturetowidth}
    \end{subfigure}
    \hfill
    \begin{subfigure}{0.24\linewidth}
        \begin{scaletikzpicturetowidth}{\textwidth}
            \begin{tikzpicture}[scale=\tikzscale]
                \begin{axis}[
                    title={CPM15},
                    xlabel={Percentage of cell annotations [\%]},
                    ylabel={DICE},
                    xmin=5, xmax=105,
                    ymin=0, ymax=0.9,
                    xtick={10,20,30,40,50,60,70,80,90,100},
                    legend pos=south east,
                    ymajorgrids=true,
                    grid style=dashed,]
                \addplot
                    coordinates {(10,0.75)(20,0.76)(30,0.79)(40,0.82)(50,0.81)(60,0.82)(70,0.8)(80,0.79)(90,0.8)(100,0.81)}; \addlegendentry{ECE (ours)}
                \addplot
                    coordinates {(10,0.33)(20,0.33)(30,0.13)(40,0.14)(50,0.16)(60,0.68)(70,0.73)(80,0.79)(90,0.79)(100,0.8)};
                    \addlegendentry{CE}
                \addplot[dashed, color=cyan, mark=otimes*]
                    coordinates {(10,0.68)(20,0.79)(30,0.76)(40,0.78)(50,0.73)(60,0.68)(70,0.71)(80,0.70)(90,0.72)(100,0.73)};
                    \addlegendentry{FL}
                \addplot[dashed, color=brown, mark=square*]
                    coordinates {(10,0.32)(20,0.31)(30,0.31)(40,0.35)(50,0.39)(60,0.41)(70,0.34)(80,0.66)(90,0.79)(100,0.75)};
                    \addlegendentry{Huber}
                \end{axis}
            \end{tikzpicture}
        \end{scaletikzpicturetowidth}
    \end{subfigure}
    \hfill
    \begin{subfigure}{0.24\linewidth}
        \begin{scaletikzpicturetowidth}{\textwidth}
            \begin{tikzpicture}[scale=\tikzscale]
                \begin{axis}[
                    title={CPM17},
                    xlabel={Percentage of cell annotations [\%]},
                    ylabel={DICE},
                    xmin=5, xmax=105,
                    xtick={10,20,30,40,50,60,70,80,90,100},
                    legend pos=south east,
                    ymajorgrids=true,
                    grid style=dashed,]
                \addplot
                    coordinates {(10,0.69)(20,0.69)(30,0.81)(40,0.85)(50,0.85)(60,0.85)(70,0.85)(80,0.84)(90,0.85)(100,0.84)}; \addlegendentry{ECE (ours)}
                \addplot
                    coordinates {(10,0.61)(20,0.62)(30,0.62)(40,0.73)(50,0.74)(60,0.79)(70,0.81)(80,0.83)(90,0.82)(100,0.83)};
                    \addlegendentry{CE}
                \addplot[dashed, color=cyan, mark=otimes*]
                    coordinates {(10,0.82)(20,0.81)(30,0.74)(40,0.76)(50,0.76)(60,0.79)(70,0.73)(80,0.73)(90,0.74)(100,0.74)};
                    \addlegendentry{FL}
                \addplot[dashed, color=brown, mark=square*]
                    coordinates {(10,0.58)(20,0.59)(30,0.60)(40,0.62)(50,0.62)(60,0.81)(70,0.81)(80,0.82)(90,0.83)(100,0.84)};
                    \addlegendentry{Huber}
                \end{axis}
            \end{tikzpicture}
        \end{scaletikzpicturetowidth}
    \end{subfigure}
    \hfill
    \begin{subfigure}{0.24\linewidth}
        \begin{scaletikzpicturetowidth}{\textwidth}
            \begin{tikzpicture}[scale=\tikzscale]
                \begin{axis}[
                    title={CRCHisto},
                    xlabel={Percentage of cell annotations [\%]},
                    ylabel={DICE},
                    xmin=5, xmax=105,
                    xtick={10,20,30,40,50,60,70,80,90,100},
                    legend pos=south east,
                    ymajorgrids=true,
                    grid style=dashed,]
                \addplot
                    coordinates {(10,0.71)(20,0.73)(30,0.73)(40,0.76)(50,0.77)(60,0.76)(70,0.76)(80,0.76)(90,0.75)(100,0.76)}; \addlegendentry{ECE (ours)}
                \addplot
                    coordinates {(10,0.39)(20,0.38)(30,0.42)(40,0.48)(50,0.55)(60,0.69)(70,0.73)(80,0.74)(90,0.75)(100,0.77)};
                    \addlegendentry{CE}
                \addplot[dashed, color=cyan, mark=otimes*]
                    coordinates {(10,0.74)(20,0.73)(30,0.68)(40,0.68)(50,0.69)(60,0.66)(70,0.67)(80,0.68)(90,0.69)(100,0.67)};
                    \addlegendentry{FL}
                \addplot[dashed, color=brown, mark=square*]
                    coordinates {(10,0.41)(20,0.41)(30,0.40)(40,0.42)(50,0.50)(60,0.62)(70,0.70)(80,0.76)(90,0.73)(100,0.73)};
                    \addlegendentry{Huber}
                \end{axis}
            \end{tikzpicture}
        \end{scaletikzpicturetowidth}
    \end{subfigure}
    \hfill
    \begin{subfigure}{0.24\linewidth}
        \begin{scaletikzpicturetowidth}{\textwidth}
            \begin{tikzpicture}[scale=\tikzscale]
                \begin{axis}[
                    title={Kumar},
                    xlabel={Percentage of cell annotations [\%]},
                    ylabel={DICE},
                    xmin=5, xmax=105,
                    xtick={10,20,30,40,50,60,70,80,90,100},
                    legend pos=south east,
                    ymajorgrids=true,
                    grid style=dashed,]
                \addplot
                    coordinates {(10,0.69)(20,0.66)(30,0.72)(40,0.80)(50,0.78)(60,0.80)(70,0.80)(80,0.81)(90,0.82)(100,0.80)}; \addlegendentry{ECE (ours)}
                \addplot
                    coordinates {(10,0.62)(20,0.62)(30,0.62)(40,0.65)(50,0.68)(60,0.77)(70,0.77)(80,0.80)(90,0.79)(100,0.81)};
                    \addlegendentry{CE}
                \addplot[dashed, color=cyan, mark=otimes*]
                    coordinates {(10,0.77)(20,0.78)(30,0.73)(40,0.71)(50,0.64)(60,0.66)(70,0.71)(80,0.73)(90,0.74)(100,0.73)};
                    \addlegendentry{FL}
                \addplot[dashed, color=brown, mark=square*]
                    coordinates {(10,0.63)(20,0.63)(30,0.63)(40,0.66)(50,0.63)(60,0.76)(70,0.79)(80,0.79)(90,0.8)(100,0.79)};
                    \addlegendentry{Huber}
                \end{axis}
            \end{tikzpicture}
        \end{scaletikzpicturetowidth}
    \end{subfigure}
    \hfill
    \begin{subfigure}{0.24\linewidth}
        \begin{scaletikzpicturetowidth}{\textwidth}
            \begin{tikzpicture}[scale=\tikzscale]
                \begin{axis}[
                    title={MoNuSeg},
                    xlabel={Percentage of cell annotations [\%]},
                    ylabel={DICE},
                    xmin=5, xmax=105,
                    ymin=0.4725, ymax=0.825,
                    xtick={10,20,30,40,50,60,70,80,90,100},
                    legend pos=south east,
                    ymajorgrids=true,
                    grid style=dashed,]
                \addplot
                    coordinates {(10,0.68)(20,0.71)(30,0.70)(40,0.69)(50,0.74)(60,0.76)(70,0.76)(80,0.78)(90,0.77)(100,0.76)}; \addlegendentry{ECE (ours)}
                \addplot
                    coordinates {(10,0.63)(20,0.63)(30,0.62)(40,0.62)(50,0.65)(60,0.77)(70,0.79)(80,0.77)(90,0.77)(100,0.80)};
                    \addlegendentry{CE}
                \addplot[dashed, color=cyan, mark=otimes*]
                    coordinates {(10,0.74)(20,0.76)(30,0.71)(40,0.68)(50,0.71)(60,0.66)(70,0.68)(80,0.62)(90,0.67)(100,0.62)};
                    \addlegendentry{FL}
                \addplot[dashed, color=brown, mark=square*]
                    coordinates {(10,0.64)(20,0.67)(30,0.64)(40,0.58)(50,0.60)(60,0.65)(70,0.69)(80,0.76)(90,0.75)(100,0.77)};
                    \addlegendentry{Huber}
                \end{axis}
            \end{tikzpicture}
        \end{scaletikzpicturetowidth}
    \end{subfigure}
    \hfill
    \begin{subfigure}{0.24\linewidth}
        \begin{scaletikzpicturetowidth}{\textwidth}
            \begin{tikzpicture}[scale=\tikzscale]
                \begin{axis}[
                    title={WBC-NuClick},
                    xlabel={Percentage of cell annotations [\%]},
                    ylabel={DICE},
                    xmin=5, xmax=105,
                    xtick={10,20,30,40,50,60,70,80,90,100},
                    legend pos=south east,
                    ymajorgrids=true,
                    grid style=dashed,]
                \addplot
                    coordinates {(10,0.86)(20,0.84)(30,0.95)(40,0.92)(50,0.95)(60,0.96)(70,0.95)(80,0.95)(90,0.94)(100,0.95)}; \addlegendentry{ECE (ours)}
                \addplot
                    coordinates {(10,0.00)(20,0.16)(30,0.31)(40,0.38)(50,0.80)(60,0.89)(70,0.93)(80,0.94)(90,0.94)(100,0.95)};
                    \addlegendentry{CE}
                \addplot[dashed, color=cyan, mark=otimes*]
                    coordinates {(10,0.40)(20,0.97)(30,0.96)(40,0.96)(50,0.96)(60,0.95)(70,0.95)(80,0.95)(90,0.95)(100,0.95)};
                    \addlegendentry{FL}
                \addplot[dashed, color=brown, mark=square*]
                    coordinates {(10,0.10)(20,0.20)(30,0.35)(40,0.62)(50,0.83)(60,0.83)(70,0.86)(80,0.95)(90,0.95)(100,0.95)};
                    \addlegendentry{Huber}
                \end{axis}
            \end{tikzpicture}
        \end{scaletikzpicturetowidth}
    \end{subfigure}
    \caption{Segmentation results on the non-exhaustive sets of the datasets}
    \label{fig:results_line_graphs_seg}
\end{figure*}
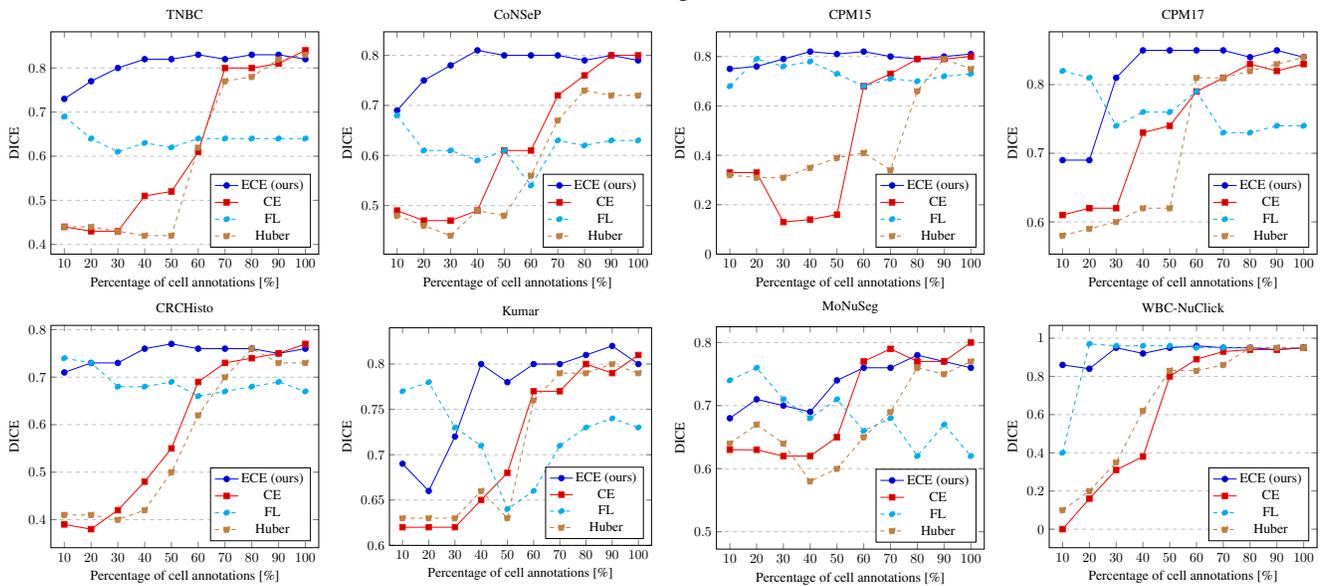

We present results for segmentation in figure~\ref{fig:results_line_graphs_seg} using standard cross-entropy (CE) (assuming what is not annotated is a negative sample), focal loss (FL) \cite{lin2017focal}, Huber loss \cite{gupta2020robust}, and exclusive cross-entropy (ECE).
The exclusive cross-entropy attains top performance in most datasets and settings.
Importantly, the exclusive cross-entropy reaches its near maximum performance consistently with only 40\% of the annotations, no matter the dataset. 
Compared to standard cross-entropy, exclusive cross-entropy improves up to 85\%, especially with sparser annotations (\eg, 10\% or 20\% variants) and harder datasets (datasets that CE scores less than 0.5 in DICE score with 10\% of annotations).
A surprising result is that focal loss achieves relative better performance in some more sparsely annotated sets, but its performance drops with more exhaustively annotated sets; and is significantly worse than the other methods. A possible cause for this finding is that the focal loss was originally designed for class object loss imbalance during detection with exhaustive annotations \cite{lin2017focal}. Therefore, the focal loss cross-entropy component, applied to all terms of the loss function, down weighs both $\cF$ and $\cFb$ groups equally.

\begin{table}
    \caption{Quantitative results on the TIL localisation dataset scored by the exclusive recall metric.}
    \vspace{-1em}
    \begin{center}
        \begin{small}
        \begin{tabular}{lcccc}
            \toprule
            & Cross-entropy & Focal loss & Huber loss & ECE\\
            \midrule
             $\text{Rec}_{\text{exc}}$ ($\uparrow$) & 0.85 & 0.81 & 0.69 & \textbf{0.88}\\
            \bottomrule
        \end{tabular}
        \end{small}
    \end{center}
    \vspace{-2em}
    \label{tab:tils-results}
\end{table}

\begin{figure*}
    \centering
    \includegraphics[width=0.75\linewidth]{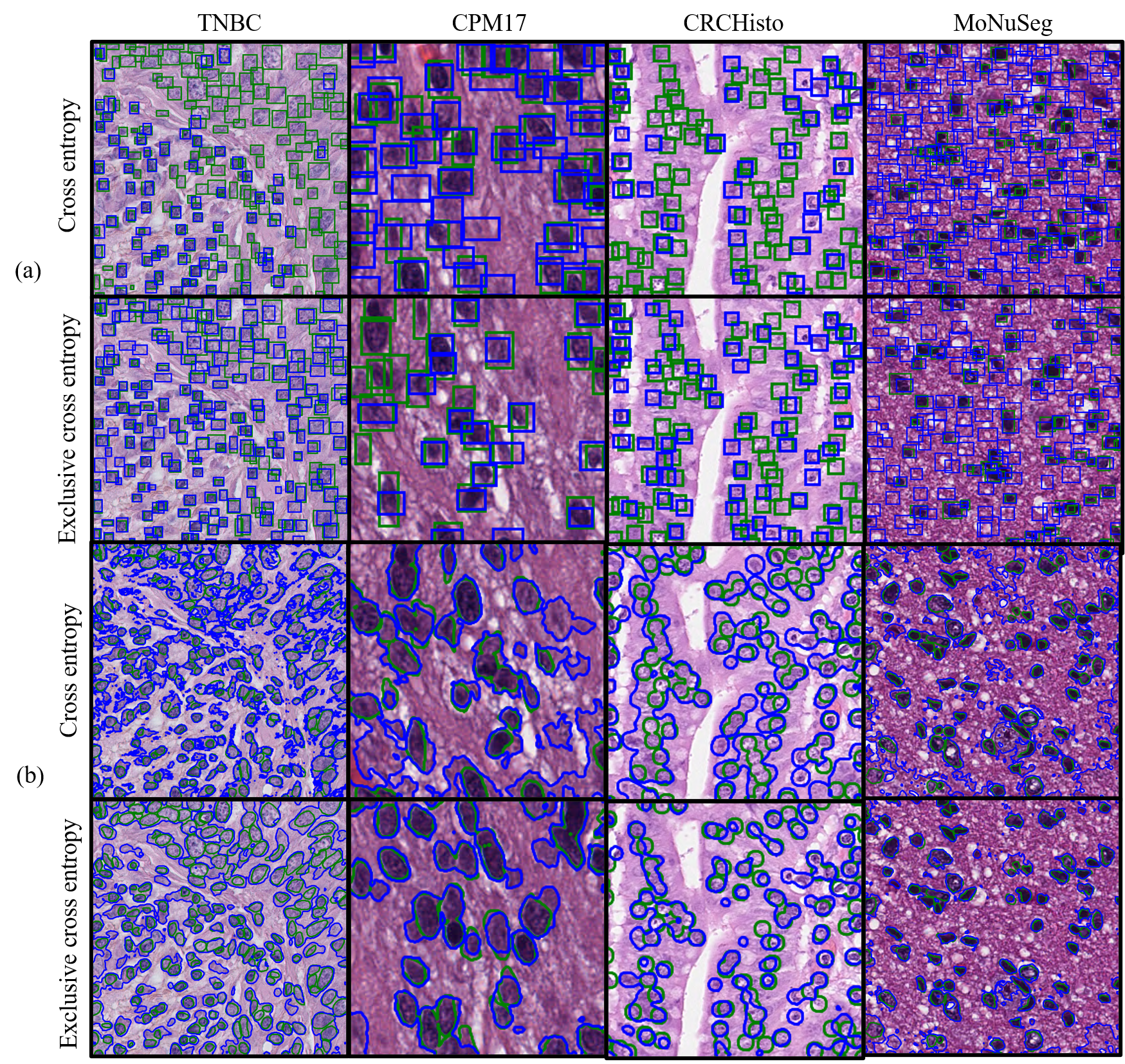}
    \caption{Qualitative results of \textbf{\textcolor{ForestGreen}{ground truth}} and \textbf{\textcolor{Blue}{predictions}} for datasets TNBC, CPM17, CRCHisto, and MoNuSeg on the 30\% non-exhaustive annotation variants for detection (a) and segmentation (b).}
    \vspace{-1em}
    \label{fig:qualitative-results-all-loses}
\end{figure*}

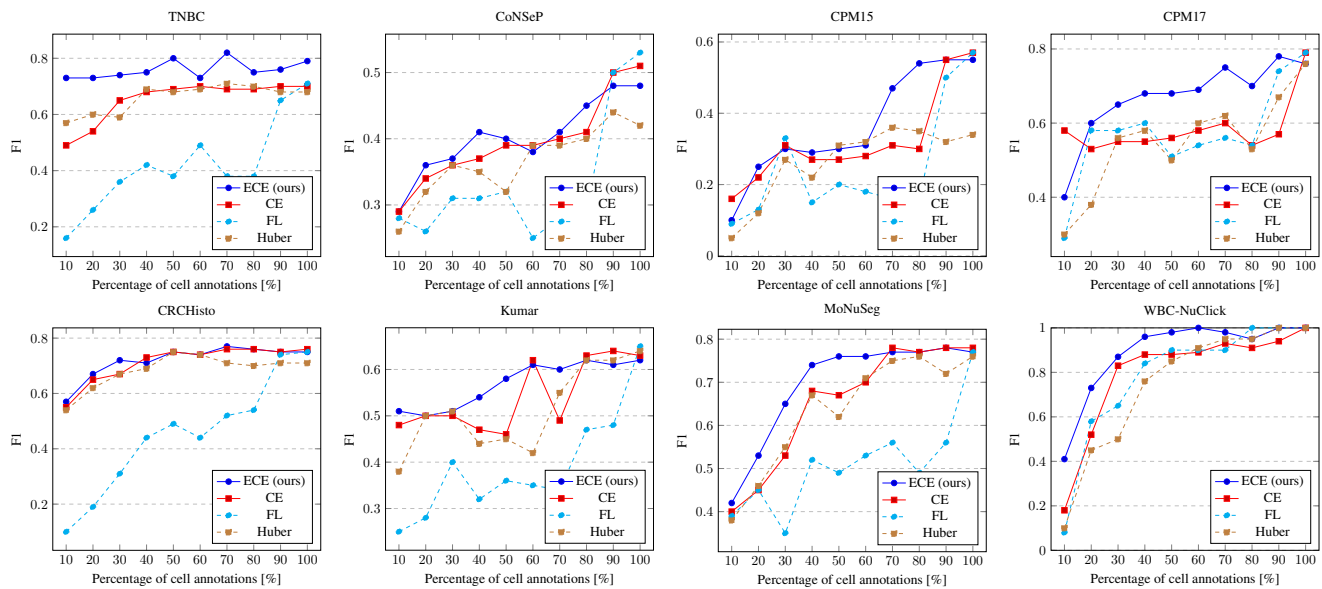
\begin{figure*}[t!]
    \centering
    \begin{subfigure}{0.24\linewidth}
        \begin{scaletikzpicturetowidth}{\textwidth}
            \begin{tikzpicture}[scale=\tikzscale]
                \begin{axis}[
                    title={TNBC},
                    xlabel={Percentage of cell annotations [\%]},
                    ylabel={F1},
                    xmin=5, xmax=105,
                    xtick={10,20,30,40,50,60,70,80,90,100},
                    legend pos=south east,
                    ymajorgrids=true,
                    grid style=dashed,]
                \addplot
                    coordinates {(10,0.73)(20,0.73)(30,0.74)(40,0.75)(50,0.80)(60,0.73)(70,0.82)(80,0.75)(90,0.76)(100,0.79)}; \addlegendentry{ECE (ours)}
                \addplot
                    coordinates {(10,0.49)(20,0.54)(30,0.65)(40,0.68)(50,0.69)(60,0.70)(70,0.69)(80,0.69)(90,0.7)(100,0.7)};
                    \addlegendentry{CE}
                \addplot[dashed, color=cyan, mark=otimes*]
                    coordinates {(10,0.16)(20,0.26)(30,0.36)(40,0.42)(50,0.38)(60,0.49)(70,0.38)(80,0.38)(90,0.65)(100,0.71)};
                    \addlegendentry{FL}
                \addplot[dashed, color=brown, mark=square*]
                    coordinates {(10,0.57)(20,0.60)(30,0.59)(40,0.69)(50,0.68)(60,0.69)(70,0.71)(80,0.70)(90,0.68)(100,0.68)};
                    \addlegendentry{Huber}
                \end{axis}
            \end{tikzpicture}
        \end{scaletikzpicturetowidth}
    \end{subfigure}
    \hfill
    \begin{subfigure}{0.24\linewidth}
        \begin{scaletikzpicturetowidth}{\textwidth}
            \begin{tikzpicture}[scale=\tikzscale]
                \begin{axis}[
                    title={CoNSeP},
                    xlabel={Percentage of cell annotations [\%]},
                    ylabel={F1},
                    xmin=5, xmax=105,
                    xtick={10,20,30,40,50,60,70,80,90,100},
                    legend pos=south east,
                    ymajorgrids=true,
                    grid style=dashed,]
                \addplot
                    coordinates {(10,0.29)(20,0.36)(30,0.37)(40,0.41)(50,0.40)(60,0.38)(70,0.41)(80,0.45)(90,0.48)(100,0.48)}; \addlegendentry{ECE (ours)}
                \addplot
                    coordinates {(10,0.29)(20,0.34)(30,0.36)(40,0.37)(50,0.39)(60,0.39)(70,0.40)(80,0.41)(90,0.5)(100,0.51)};
                    \addlegendentry{CE}
                \addplot[dashed, color=cyan, mark=otimes*]
                    coordinates {(10,0.28)(20,0.26)(30,0.31)(40,0.31)(50,0.32)(60,0.25)(70,0.28)(80,0.29)(90,0.5)(100,0.53)};
                    \addlegendentry{FL}
                \addplot[dashed, color=brown, mark=square*]
                    coordinates {(10,0.26)(20,0.32)(30,0.36)(40,0.35)(50,0.32)(60,0.39)(70,0.39)(80,0.40)(90,0.44)(100,0.42)};
                    \addlegendentry{Huber}
                \end{axis}
            \end{tikzpicture}
        \end{scaletikzpicturetowidth}
    \end{subfigure}
    \hfill
    \begin{subfigure}{0.24\linewidth}
        \begin{scaletikzpicturetowidth}{\textwidth}
            \begin{tikzpicture}[scale=\tikzscale]
                \begin{axis}[
                    title={CPM15},
                    xlabel={Percentage of cell annotations [\%]},
                    ylabel={F1},
                    xmin=5, xmax=105,
                    xtick={10,20,30,40,50,60,70,80,90,100},
                    legend pos=south east,
                    ymajorgrids=true,
                    grid style=dashed,]
                \addplot
                    coordinates {(10,0.10)(20,0.25)(30,0.30)(40,0.29)(50,0.30)(60,0.31)(70,0.47)(80,0.54)(90,0.55)(100,0.55)}; \addlegendentry{ECE (ours)}
                \addplot
                    coordinates {(10,0.16)(20,0.22)(30,0.31)(40,0.27)(50,0.27)(60,0.28)(70,0.31)(80,0.30)(90,0.55)(100,0.57)};
                    \addlegendentry{CE}
                \addplot[dashed, color=cyan, mark=otimes*]
                    coordinates {(10,0.09)(20,0.13)(30,0.33)(40,0.15)(50,0.20)(60,0.18)(70,0.16)(80,0.16)(90,0.5)(100,0.57)};
                    \addlegendentry{FL}
                \addplot[dashed, color=brown, mark=square*]
                    coordinates {(10,0.05)(20,0.12)(30,0.27)(40,0.22)(50,0.31)(60,0.32)(70,0.36)(80,0.35)(90,0.32)(100,0.34)};
                    \addlegendentry{Huber}
                \end{axis}
            \end{tikzpicture}
        \end{scaletikzpicturetowidth}
    \end{subfigure}
    \hfill
    \begin{subfigure}{0.24\linewidth}
        \begin{scaletikzpicturetowidth}{\textwidth}
            \begin{tikzpicture}[scale=\tikzscale]
                \begin{axis}[
                    title={CPM17},
                    xlabel={Percentage of cell annotations [\%]},
                    ylabel={F1},
                    xmin=5, xmax=105,
                    xtick={10,20,30,40,50,60,70,80,90,100},
                    legend pos=south east,
                    ymajorgrids=true,
                    grid style=dashed,]
                \addplot
                    coordinates {(10,0.40)(20,0.60)(30,0.65)(40,0.68)(50,0.68)(60,0.69)(70,0.75)(80,0.70)(90,0.78)(100,0.76)}; \addlegendentry{ECE (ours)}
                \addplot
                    coordinates {(10,0.58)(20,0.53)(30,0.55)(40,0.55)(50,0.56)(60,0.58)(70,0.60)(80,0.54)(90,0.57)(100,0.79)};
                    \addlegendentry{CE}
                \addplot[dashed, color=cyan, mark=otimes*]
                    coordinates 
                    {(10,0.29)(20,0.58)(30,0.58)(40,0.60)(50,0.51)(60,0.54)(70,0.56)(80,0.54)(90,0.74)(100,0.79)};
                    \addlegendentry{FL}
                \addplot[dashed, color=brown, mark=square*]
                    coordinates {(10,0.30)(20,0.38)(30,0.56)(40,0.58)(50,0.50)(60,0.60)(70,0.62)(80,0.53)(90,0.67)(100,0.76)};
                    \addlegendentry{Huber}
                \end{axis}
            \end{tikzpicture}
        \end{scaletikzpicturetowidth}
    \end{subfigure}
    \hfill
    \begin{subfigure}{0.24\linewidth}
        \begin{scaletikzpicturetowidth}{\textwidth}
            \begin{tikzpicture}[scale=\tikzscale]
                \begin{axis}[
                    title={CRCHisto},
                    xlabel={Percentage of cell annotations [\%]},
                    ylabel={F1},
                    xmin=5, xmax=105,
                    xtick={10,20,30,40,50,60,70,80,90,100},
                    legend pos=south east,
                    ymajorgrids=true,
                    grid style=dashed,]
                \addplot
                    coordinates {(10,0.57)(20,0.67)(30,0.72)(40,0.71)(50,0.75)(60,0.74)(70,0.77)(80,0.76)(90,0.75)(100,0.75)}; \addlegendentry{ECE (ours)}
                \addplot
                    coordinates {(10,0.55)(20,0.65)(30,0.67)(40,0.73)(50,0.75)(60,0.74)(70,0.76)(80,0.76)(90,0.75)(100,0.76)};
                    \addlegendentry{CE}
                \addplot[dashed, color=cyan, mark=otimes*]
                    coordinates {(10,0.10)(20,0.19)(30,0.31)(40,0.44)(50,0.49)(60,0.44)(70,0.52)(80,0.54)(90,0.74)(100,0.75)};
                    \addlegendentry{FL}
                \addplot[dashed, color=brown, mark=square*]
                    coordinates {(10,0.54)(20,0.62)(30,0.67)(40,0.69)(50,0.75)(60,0.74)(70,0.71)(80,0.70)(90,0.71)(100,0.71)};
                    \addlegendentry{Huber}
                \end{axis}
            \end{tikzpicture}
        \end{scaletikzpicturetowidth}
    \end{subfigure}
    \hfill
    \begin{subfigure}{0.24\linewidth}
        \begin{scaletikzpicturetowidth}{\textwidth}
            \begin{tikzpicture}[scale=\tikzscale]
                \begin{axis}[
                    title={Kumar},
                    xlabel={Percentage of cell annotations [\%]},
                    ylabel={F1},
                    xmin=5, xmax=105,
                    xtick={10,20,30,40,50,60,70,80,90,100},
                    legend pos=south east,
                    ymajorgrids=true,
                    grid style=dashed,]
                \addplot
                    coordinates {(10,0.51)(20,0.50)(30,0.51)(40,0.54)(50,0.58)(60,0.61)(70,0.60)(80,0.62)(90,0.61)(100,0.62)}; \addlegendentry{ECE (ours)}
                \addplot
                    coordinates {(10,0.48)(20,0.50)(30,0.50)(40,0.47)(50,0.46)(60,0.62)(70,0.49)(80,0.63)(90,0.64)(100,0.63)};
                    \addlegendentry{CE}
                \addplot[dashed, color=cyan, mark=otimes*]
                    coordinates {(10,0.25)(20,0.28)(30,0.40)(40,0.32)(50,0.36)(60,0.35)(70,0.34)(80,0.47)(90,0.48)(100,0.65)};
                    \addlegendentry{FL}
                \addplot[dashed, color=brown, mark=square*]
                    coordinates {(10,0.38)(20,0.50)(30,0.51)(40,0.44)(50,0.45)(60,0.42)(70,0.55)(80,0.62)(90,0.62)(100,0.64)};
                    \addlegendentry{Huber}
                \end{axis}
            \end{tikzpicture}
        \end{scaletikzpicturetowidth}
    \end{subfigure}
    \hfill
    \begin{subfigure}{0.24\linewidth}
        \begin{scaletikzpicturetowidth}{\textwidth}
            \begin{tikzpicture}[scale=\tikzscale]
                \begin{axis}[
                    title={MoNuSeg},
                    xlabel={Percentage of cell annotations [\%]},
                    ylabel={F1},
                    xmin=5, xmax=105,
                    xtick={10,20,30,40,50,60,70,80,90,100},
                    legend pos=south east,
                    ymajorgrids=true,
                    grid style=dashed,]
                \addplot
                    coordinates {(10,0.42)(20,0.53)(30,0.65)(40,0.74)(50,0.76)(60,0.76)(70,0.77)(80,0.77)(90,0.78)(100,0.77)}; \addlegendentry{ECE (ours)}
                \addplot
                    coordinates {(10,0.40)(20,0.45)(30,0.53)(40,0.68)(50,0.67)(60,0.70)(70,0.78)(80,0.77)(90,0.78)(100,0.78)};
                    \addlegendentry{CE}
                \addplot[dashed, color=cyan, mark=otimes*]
                    coordinates {(10,0.39)(20,0.45)(30,0.35)(40,0.52)(50,0.49)(60,0.53)(70,0.56)(80,0.49)(90,0.56)(100,0.77)};
                    \addlegendentry{FL}
                \addplot[dashed, color=brown, mark=square*]
                    coordinates {(10,0.38)(20,0.46)(30,0.55)(40,0.67)(50,0.62)(60,0.71)(70,0.75)(80,0.76)(90,0.72)(100,0.76)};
                    \addlegendentry{Huber}
                \end{axis}
            \end{tikzpicture}
        \end{scaletikzpicturetowidth}
    \end{subfigure}
    \hfill
    \begin{subfigure}{0.24\linewidth}
        \begin{scaletikzpicturetowidth}{\textwidth}
            \begin{tikzpicture}[scale=\tikzscale]
                \begin{axis}[
                    title={WBC-NuClick},
                    xlabel={Percentage of cell annotations [\%]},
                    ylabel={F1},
                    xmin=5, xmax=105,
                    ymin=0, ymax=1,
                    xtick={10,20,30,40,50,60,70,80,90,100},
                    legend pos=south east,
                    ymajorgrids=true,
                    grid style=dashed,]
                \addplot
                    coordinates {(10,0.41)(20,0.73)(30,0.87)(40,0.96)(50,0.98)(60,1)(70,0.98)(80,0.95)(90,1)(100,1)}; \addlegendentry{ECE (ours)}
                \addplot
                    coordinates {(10,0.18)(20,0.52)(30,0.83)(40,0.88)(50,0.88)(60,0.89)(70,0.93)(80,0.91)(90,0.94)(100,1)};
                    \addlegendentry{CE}
                \addplot[dashed, color=cyan, mark=otimes*]
                    coordinates {(10,0.08)(20,0.58)(30,0.65)(40,0.84)(50,0.90)(60,0.90)(70,0.90)(80,1)(90,1)(100,1)};
                    \addlegendentry{FL}
                \addplot[dashed, color=brown, mark=square*]
                    coordinates {(10,0.10)(20,0.45)(30,0.50)(40,0.76)(50,0.85)(60,0.91)(70,0.95)(80,0.95)(90,1)(100,1)};
                    \addlegendentry{Huber}
                \end{axis}
            \end{tikzpicture}
        \end{scaletikzpicturetowidth}
    \end{subfigure}
    \caption{Detection results on the non-exhaustive sets of the datasets}
    \label{fig:results_line_graphs_det}
\end{figure*}
\subsection{Sparse-shot detection}

We present results for box detection in figure~\ref{fig:results_line_graphs_det} with the same hyper-parameters as in segmentation.
In MoNuSeg, WBC-Nuclick and TNBC exclusive cross-entropy loss outperforms the standard cross-entropy and the focal loss consistently by about 10\% in the 10-50\% variants.
In CRCHisto, CoNSeP, Kumar and CPM15, exclusive cross-entropy still maintains top performance, but for different non-exhaustive variants it is matched by different methods, showing more robustness in the final predictions.
A possible reason for the smaller increase of performance compared to segmentation is that segmentation is more challenging than detection. This can be due to the fact that the number of output objects, pixels, in segmentation is larger than the number of objects, cells, in detection. Hence, the number of unannotated objects is relatively lower in the detection task.
Last, focal loss appears to have trouble with balancing between the foreground and background due to its uniform weighing strategy.

We, furthermore, present results in terms of exclusive recall on the TIL dataset in table \ref{tab:tils-results}.
Exclusive cross-entropy performs best, locating correctly the most true positive lymphocytes, while not confusing them with other visually similar cell types like tumour cells or fibroblasts. Upon visual inspection, the difference between the methods is even greater but not quantitatively reflected due to the large number of missing annotations; as discussed in the supplementary material.

\subsection{Qualitative results}

We show in figure \ref{fig:qualitative-results-all-loses} qualitative results for cross-entropy and exclusive cross-entropy.
Cross-entropy tends to either under-predict, mostly in detection, or over-predict in segmentation. 
Exclusive cross-entropy correctly detects most objects while avoiding erroneous background predictions.


\section{Conclusion}

In this work, we focus on the problem of \emph{sparse-shot learning}, especially in the context of localising extremely many objects.
Sparse-shot learning is particularly important for certain types of images, like digitised tissue sections in computational pathology, easily exceeding resolutions of 250'000 $\times$ 250'000 pixels and millions of cells to be localised.
We show that standard cross-entropy assuming all background as negative labels leads to biased learning and poor optimisation, likely due to the contributions represented by large second-order derivatives in the loss.
By ignoring these terms, we present \emph{exclusive cross-entropy}.
Extensive experiments on nine datasets and two localisation tasks, detection with YOLLO and segmentation with Unet, show that we obtain considerable improvements compared to cross-entropy or focal loss, while often reaching the best possible accuracy for the model with only 10-40\% of annotations present.
 



\section*{Acknowledgements}
The collaboration project is co-funded by the PPP Allowance made available by Health Holland\footnote{\href{https://www.health-holland.com}{https://www.health-holland.com}}. Top Sector Life Sciences $\&$ Health, to stimulate public-private partnerships.

{\small
\bibliographystyle{unsrt}
\bibliography{bib.bib}
}

\end{document}